%% file: root.tex
\documentclass[letterpaper, 10 pt, conference]{ieeeconf}  

\IEEEoverridecommandlockouts                              

\usepackage{url} 
\usepackage{algorithm}
\usepackage{algpseudocode}
\usepackage{array} 
\usepackage{graphicx}
\usepackage{amsmath} 
\usepackage{bm}
\usepackage{comment}
\usepackage{booktabs}       
\usepackage{multirow}
\usepackage{makecell}
\usepackage{adjustbox}
\usepackage{xcolor}
\usepackage{cite}
\usepackage{hyperref}

\usepackage{bbold}
\usepackage{amsfonts}
\usepackage{amssymb}
\usepackage{txfonts}

\usepackage{booktabs}
\usepackage{nicematrix}
\usepackage{tcolorbox}
\tcbuselibrary{breakable}

\newcolumntype{C}{>{\centering\arraybackslash}p{4.5em}}

\title{\Large \bf
PROGrasp: Pragmatic Human-Robot Communication for Object Grasping 
}
\author{Gi-Cheon Kang$^{1,3}$ \,\, Junghyun Kim$^{1,3}$ \,\, Jaein Kim$^{2,3}$ \,\, Byoung-Tak Zhang$^{1,2,3}$%
\thanks{$^{1}$Interdisciplinary Program in AI, Seoul National University}
\thanks{$^{2}$Interdisciplinary Program in Neuroscience, Seoul National University}
\thanks{$^{3}$AI Institute, Seoul National University}
\thanks{This research was supported by the Institute of Information and Communications Technology Planning and Evaluation (IITP) (2021-0-01343-GSAI/40\%, 2022-0-00953-PICA/30\%, 2022-0-00951-LBA/10\%, 2021-0-02068-AIHub/10\%) and National Research Foundation of Korea (NRF) (RS-2023-00274280/10\%) grant funded by the Korean government.}
}

\begin{document}

\maketitle
\thispagestyle{empty}
\pagestyle{empty}
\begin{abstract}
Interactive Object Grasping (IOG) is the task of identifying and grasping the desired object via human-robot natural language interaction. Current IOG systems assume that a human user initially specifies the target object's category (\textit{e.g.,} bottle). Inspired by \textit{pragmatics}, where humans often convey their intentions by relying on context to achieve goals, we introduce a new IOG task, Pragmatic-IOG, and the corresponding dataset, Intention-oriented Multi-modal Dialogue (IM-Dial). In our proposed task scenario, an intention-oriented utterance (\textit{e.g.,} ``I am thirsty'') is initially given to the robot. The robot should then identify the target object by interacting with a human user. Based on the task setup, we propose a new robotic system that can interpret the user's intention and pick up the target object, Pragmatic Object Grasping (PROGrasp). PROGrasp performs Pragmatic-IOG by incorporating modules for visual grounding, question asking, object grasping, and most importantly, answer interpretation for pragmatic inference. Experimental results show that PROGrasp is effective in offline (\textit{i.e.,} target object discovery) and online (\textit{i.e.,} IOG with a physical robot arm) settings. Code and data are available at \url{https://github.com/gicheonkang/prograsp}.

\end{abstract}
\input{01_intro}

\input{02_related_works}
\input{03_method}
\input{04_experiment}
\input{05_conclusion}




\bibliographystyle{IEEEtran.bst}
\bibliography{ref}

\clearpage
\input{06_appendix}

\end{document}

%% file: 01_intro.tex
\section{INTRODUCTION}
Recent advances in robotics and artificial intelligence (AI) have made intelligent robots ubiquitous in our daily lives. To get closer to non-expert users, robots should communicate with humans using natural language and make decisions based on the interaction. Notably, in the field of human-robot interaction, there have been extensive studies~\cite{ivgorefhri, iprwowusli, invigorate, yang2022interactive, mo2022towards} on developing such robots under the umbrella of Interactive Object Grasping (IOG). A typical scenario of IOG starts mentioning the target object, such as ``Give me the plastic bottle'', but there is more than one object in the scene that meets the instruction. The robot should disambiguate the target object by asking questions to the conversational partner and then perform object grasping.

While the progress of IOG is exciting, the current scenario limits the ability for robots to understand beyond the literal meaning of natural language instructions. Specifically, instructions in the existing scenario clearly specify the category of the target object (\textit{e.g.,} bottle). In other words, current IOG systems may work properly when the target object's category is given. However, we humans often convey our \textit{intended meanings} by relying on context to achieve communicative goals~\cite{searle1969speech, frank2012predicting}. For example, when we need some water and want our conversational partner to bring it, we briefly say ``I am thirsty.'' The partner then enriches the literal meaning of the utterance based on various types of shared context (\textit{e.g.,} visual information and dialogue context) and reason about context-appropriate behavior. This ability to interpret and use language in context to achieve goals is known as pragmatics~\cite{fried2022pragmatics,goodman2016pragmatic,smith2013learning}.  
\begin{figure}[t!]
\centering
\includegraphics[width=0.48\textwidth]{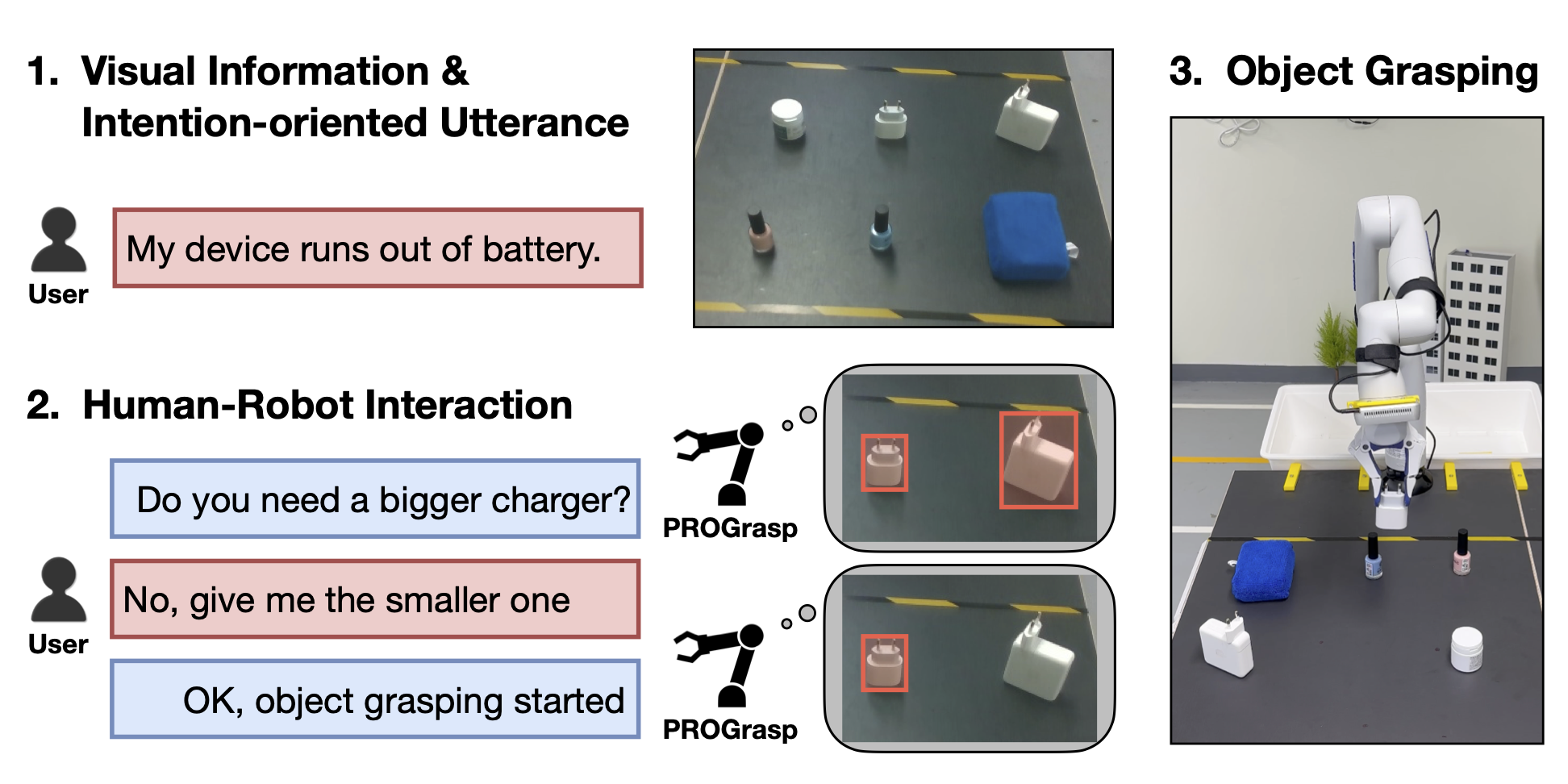}
\caption{\textbf{Overview of interactive object grasping with intention-oriented utterance.} The initial utterance does not contain the target object's category.}
\label{fig:teaser}
\vspace*{-0.5cm}
\end{figure}

We argue that the next-generation robotic system should have pragmatic reasoning ability -- capture the user's intention with contextual information and achieve the desired goal. Therefore, we introduce a new task, Pragmatic-IOG, to study pragmatic reasoning behavior in IOG. As shown in Figure~\hyperref[fig:teaser]{1}, we consider a scenario where a human user begins a conversation with an \textit{intention-oriented} utterance like ``My device runs out of battery.'' The robot should then find all valid object candidates (\textit{e.g.,} the red-colored object regions in Figure~\hyperref[fig:teaser]{1}) via visual grounding~\cite{yu2016modeling} and ask a question for disambiguation. After receiving the user's response, the robot pinpoints the target object and grasps the desired object. To study the problem, we propose a new dataset, called Intention-oriented Multi-modal Dialogue (IM-Dial). The IM-Dial dataset contains 800 images and 500 human-to-human dialogue data regarding 86 categories of everyday objects. The dialogue consists of the intention-oriented utterance and a series of question-and-answer pairs.

We propose a new robotic system that can reason about intention-oriented natural language utterances of the user and grasp desired objects through human-robot interaction, called \textbf{PR}agmatic \textbf{O}bject \textbf{Grasp}ing (PROGrasp). PROGrasp consists of four modules: (1) a visual grounding module (VG) that predicts the region coordinates of valid objects, (2) a question generation module (Q-gen) that learns to generate questions to identify the user's intention, (3) an answer interpretation module (A-int) that interprets the human user's response given the multi-modal context, and (4) an object grasping module (OG) to pick up the inferred object. 

PROGrasp trains VG, Q-gen, and A-int on the IM-Dial data. After training, PROGrasp performs our proposed task by interacting with the human user. Specifically, VG first predicts a set of object region candidates given the intention-oriented utterance. Q-gen then generates a question, and the user responds to the question. Next, PROGrasp determines the target region among the region candidates based on how well each candidate region explains the visual and dialogue context, which we call \textit{pragmatic inference}. We implement pragmatic inference as a multi-agent reasoning of VG and A-int where VG evaluates the likelihood of each region candidate given the visual and dialogue context, and A-int rescores alternative region candidates by interpreting the user's response. Finally, OG computes the 3D coordinates of the inferred region and performs object grasping.

We conduct offline and online experiments on the IM-Dial dataset. In offline experiments, we study how well PROGrasp identifies the target object. PROGrasp significantly improves the accuracy of offline experiments by 35\% compared with the baselines. Moreover, PROGrasp outperforms the powerful multimodal foundation model~\cite{gpt4v} on validation data. In online experiments, we use a physical robot arm to evaluate the success rate of object grasping. PROGrasp boosts the success rate by 17\%. Furthermore, our system efficiently identifies the target object through fewer interactions than baselines. Finally, we perform qualitative analysis, visualizing diverse samples inferred by PROGrasp.

Our contributions are three-fold. First, we propose an interactive object-grasping system (\textit{i.e.,} PROGrasp) that capably understands the human user's intention and grasps the desired object through dialogue. Second, we introduce a new task Pragmatic-IOG with a novel dataset, Intention-oriented Multi-modal Dialogue (IM-Dial). Third, through extensive experiments, our robotic system validates its (1) efficacy in both offline and online experiments and (2) efficiency when identifying the target object via pragmatic inference.

%% file: 02_related_works.tex
\section{RELATED WORK}

\noindent\textbf{Language-Guided Object Grasping.} There has been extensive research on developing object-grasping systems that can understand natural language. Some studies~\cite{paul2016efficient,shridhar2017grounding,venkatesh2021spatial,rorwcnlq,kim2023gvcci} make robots manipulate objects only with initial language instruction, assuming that the instruction is enough to identify the desired object. However, natural language is inherently ambiguous~\cite{piantadosi2012communicative}. Therefore, a line of research~\cite{ivgorefhri, iprwowusli, invigorate, yang2022interactive, mo2022towards}, which we call Interactive Object Grasping (IOG), considers the scenario where robots need more information to disambiguate the target object. They typically generate questions and perform object grasping based on the response from a human user. Our approach belongs to IOG, but it differs from previous studies in two aspects. First, prior work in IOG considers a scenario where the category of the target object is clearly specified. However, we design a task scenario that requires robots to understand the semantic meaning and, by extension, \textit{the intended meaning} of the user's utterance. Accordingly, intention-oriented utterances focus on the user's intention without specifying the category of the target object. Second, previous studies define the format of either questions or responses. Specifically, the questions are fixed (\textit{e.g.,} ``Which one?'')~\cite{iprwowusli,whitney2017reducing} or based on templates~\cite{ivgorefhri, invigorate, yang2022interactive, mo2022towards}. The answer formats are also binary~\cite{iprwowusli}, a single word~\cite{whitney2017reducing}, or based on a pre-defined pool~\cite{ivgorefhri, invigorate, yang2022interactive, mo2022towards}. However, PROGrasp does not impose any constraints on the format of the questions and responses. Our Q-gen generates unconstrained questions without relying on any templates, and A-int understands various types of responses, enabling non-expert users to interact with the robot more naturally. 

\noindent\textbf{Pragmatics.} There is a long history of research studying how linguistic meaning is affected by context~\cite{grice1975logic, searle1969speech, frank2012predicting} under the name of pragmatics. According to the work~\cite{fried2022pragmatics}, there are four kinds of well-studied tasks in the field of pragmatics: reference games~\cite{frank2012predicting,monroe2017colors}, image captioning~\cite{andreas2016reasoning,cohn-gordon2018pragmatically}, instruction following~\cite{chen2011learning,anderson2018vision}, and grounded dialogue~\cite{de2017guesswhat,kim2019codraw,chai2014collaborative,kang2023dialog}. Our work belongs to the last category, but it is the first work that integrates grounded goal-oriented dialogue into a real-world robot arm for object grasping. Regarding computational modeling, PROGrasp shares the spirit with the Rational Speech Acts (RSA)~\cite{frank2012predicting,goodman2016pragmatic}. We propose a multi-agent reasoning method (\textit{i.e.,} pragmatic inference) to identify target objects accurately and efficiently by interpreting the human user's response. 

%% file: 03_method.tex
\section{METHOD}

\subsection{Background}
As the Web-scale data sources~\cite{changpinyo2021conceptual,schuhmann2021laion} are publicly available, finetuning the model pre-trained on such datasets to the specific task has become a de facto standard strategy in AI. Accordingly, there has been a lot of multi-modal pre-training methods~\cite{wang2022ofa,lu2019vilbert,tan2019lxmert} trained on the large-scale image-text pairs. We employ a simple yet powerful multi-modal sequence-to-sequence model, OFA~\cite{wang2022ofa}, since it can cover various multi-modal tasks with a unified architecture. OFA is pre-trained on a wide range of multi-modal and uni-modal datasets with sequence-to-sequence learning~\cite{sutskever2014sequence,vaswani2017attention}. Specifically, the learning objective of OFA is to optimize $\mathrm{max} \sum^{|y|}_{i=1} \mathrm{log}P_\theta (y_i| y_{<i}, v, x)$, where $x$ and $y$ denote the input and target sequences, respectively. $y_{<i}$ denotes all tokens before the $i$-th token in the target sequence. $\theta$ is the model parameters, and $v$ is visual information. The encoder encodes $x$ and $v$ and conveys the hidden states to the decoder. The decoder predicts the next token $y_i$ given a set of preceding tokens $y_{<i}$ and the hidden states of the encoder. We train our proposed modules in PROGrasp (\textit{i.e.,} VG, Q-gen, and A-int) by finetuning OFA on the IM-Dial dataset. More details can be found in the following Section.

\begin{figure*}[t!]
\centering
\label{fig:prograsp}
\includegraphics[width=0.93\textwidth]{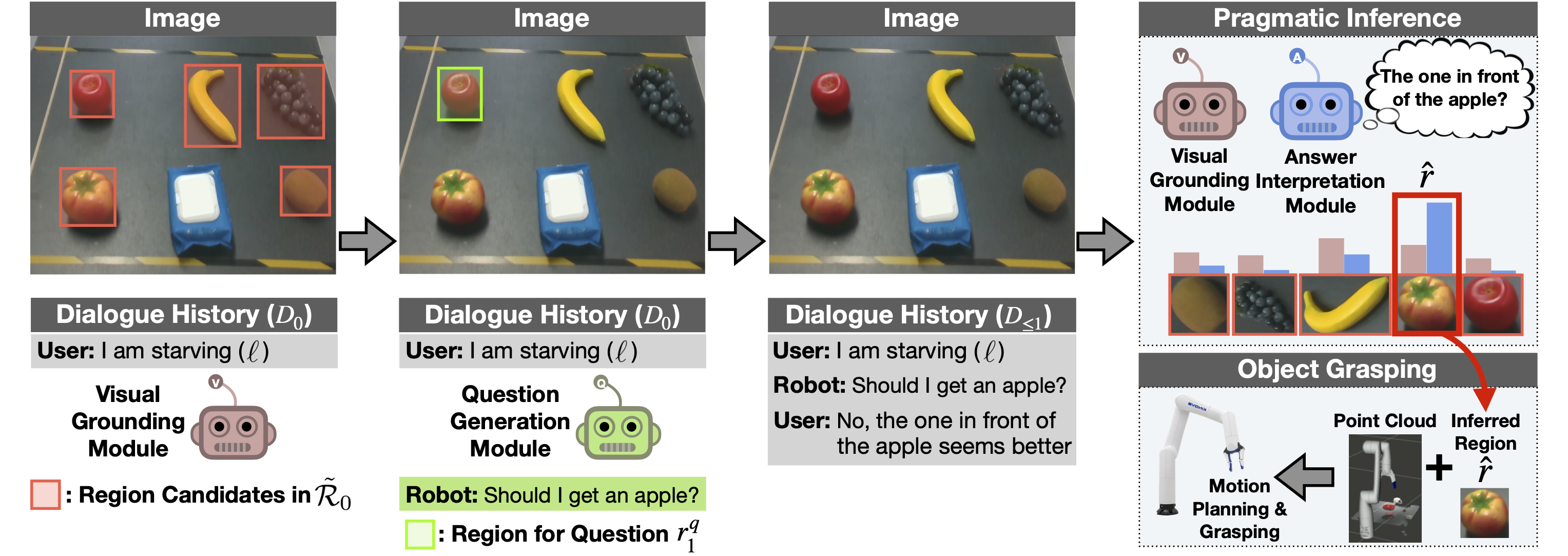}
\vspace*{-0.2cm}
\caption{\textbf{Illustration of the inference step in PROGrasp when $\bm{T=1}$.} VG first performs object grounding using the dialogue history. Q-gen then selects the object candidate to ask and generates a question. After obtaining the response from the user, VG cooperates with A-int to determine the target object region. The object grasping module finally grasps the object by computing the 3D coordinates of the target object.}
\vspace*{-0.5cm}
\end{figure*}

\subsection{Problem Statement}
The goal of Pragmatic-IOG is to discover the target object's region coordinates $r^* = \left\langle x_1, y_1, x_2, y_2 \right\rangle$ through human-robot interaction and pick up the object. We assume that a human user initially provides intention-oriented natural language utterance $\ell$, and there is one target object in the given 2D image $\mathcal{I}$. Our system asks natural language question $q$ to the user to identify the target object, and the user gives an answer $a$. The dialogue history $\mathcal{D}$ is initialized to $\ell$, and question and answer pairs at each round are appended to the dialogue history. Our robotic system predicts the region coordinates $\hat{r}$ after $T$ rounds of dialogue. Finally, it performs object grasping, computing the 3D coordinates of the target object based on $\hat{r}$ and the point cloud.

\subsection{Visual Grounding Module}
The visual grounding module (VG) aims to localize objects based on the multi-modal context. Specifically, the Intention-oriented Multi-modal Dialogue (IM-Dial) dataset contains an image $\mathcal{I}$, a visually-grounded dialogue $\mathcal{D} = \{\underbrace{\ell}_{\mathcal{D}_{0}}, \underbrace{(q_{1}, a_{1})}_{\mathcal{D}_{1}}, \cdots, \underbrace{(q_{N}, a_{N})}_{\mathcal{D}_{N}}\}$, and region labels $\mathcal{R} = \{\mathcal{R}_0, \cdots, \mathcal{R}_N\}$ where each element $\mathcal{R}_n = \{r^{vg}_i\}_{i=1}^{|\mathcal{R}_n|}$ is a set of object regions that VG should predict given the dialogue history $\mathcal{D}_{\le n}$. $\mathcal{D}_{\le n}$ denotes all dialogue data before or equal to the $n$-th round. Consequently, VG is trained to maximize the log-likelihood of the ground-truth regions. Formally, $\mathrm{max} \sum^{N}_{n=0} \mathrm{log}P_{\mathcal{V}} (\mathcal{R}_n | \mathcal{I}, \mathcal{D}_{\le n})$. PROGrasp regards the concatenation of the region coordinates in $\mathcal{R}_n$ as the target sequence in sequence-to-sequence learning and trains VG on top of the pre-trained OFA model~\cite{wang2022ofa}. After training, as shown in Figure~\hyperref[fig:prograsp]{2}, VG can predict $\tilde{\mathcal{R}}_0$ (\textit{i.e.,} the red-colored object regions) given $\mathcal{D}_0$ (\textit{i.e.,} ``I am starving.'') and an image $\mathcal{I}$.

\subsection{Question Generation \& Answer Interpretation Modules}
PROGrasp trains the question generation (Q-gen) and answer interpretation (A-int) modules to produce the utterances from the human questioner and the human answerer, respectively. First, Q-gen learns to generate human-annotated questions in the IM-Dial dataset, given an input image, the dialogue history, and an object region for questioning as follows: $\mathrm{max} \sum^{N}_{n=1} \mathrm{log}P_{\mathcal{Q}} (q_n | \mathcal{I}, \mathcal{D}_{\le n-1}, r^q_n)$. The object region $r^q_n$ is annotated by the human questioner when collecting the IM-Dial dataset. Regarding sequence-to-sequence learning, the target sequence is the question $q_n$, and the triplet $(\mathcal{I}, \mathcal{D}_{\le n-1}, r^q_n)$ is fed into the encoder. Next, the answer interpretation module (A-int), which is a proxy for the human user, learns to generate the response of the human answerer. It is optimized by maximizing the log-likelihood of the ground-truth answer: $\mathrm{max} \sum^{N}_{n=1} \mathrm{log}P_{\mathcal{A}} (a_n | \mathcal{I}, r^*, q_n)$. Note that A-int takes the ground-truth region coordinates $r^*$ during training and implicitly learns the semantic alignment between $r^*$ and the question-and-answer pair $(q_n, a_n)$. Moreover, we assume that the answer distribution is independent of the dialogue history (\textit{i.e.,} $P_{\mathcal{A}} (a_n | \mathcal{I}, \mathcal{D}_{\le n-1}, r^*, q_n) = P_{\mathcal{A}} (a_n | \mathcal{I}, r^*, q_n)$) by following the work~\cite{lee2018answerer}. We also implement both Q-gen and A-int by finetuning the pre-trained OFA model~\cite{wang2022ofa}.

\subsection{Inference Step}
The inference step of PROGrasp is described in Algorithm~\hyperref[alg:prograsp]{1} and Figure~\hyperref[fig:prograsp]{2}. PROGrasp obtains an image $\mathcal{I}$ from a camera. A human user provides intention-oriented utterance $\ell$. PROGrasp proceeds $T$ rounds of dialogue in the inference step, and $T$ is a hyperparameter. VG (\textit{i.e,} $P_\mathcal{V}$) first predicts a set of object regions $\tilde{\mathcal{R}}$ and saves it to the superset $\mathcal{R}$. PROGrasp accumulates the object region candidates predicted in each round of dialogue to maximize the probability of having the target object's region coordinates in the set $\mathcal{R}$. PROGrasp then samples an object region $r^q_t$ from $\mathcal{R}$. The image, the dialogue history, and the sampled object region are fed into Q-gen (\textit{i.e.,} $P_\mathcal{Q}$) to produce a question (\textit{e.g.,} ``Should I get a banana?'' in Figure~\hyperref[fig:prograsp]{2}). Next, the human user answers the question by checking whether the object mentioned in the question corresponds to the target object. After receiving the user's response, PROGrasp saves the question-and-answer pairs to the dialogue history and evaluates each object region candidate $r$ in the set $\mathcal{R}$. In the evaluation, VG computes the likelihood of each region candidate given the multi-modal context, and A-int rescores each candidate $r$ whether it describes the question and the user's response $(\tilde{q}_t, \tilde{a}_t)$. In other words, VG cooperates with A-int to determine the best region $\hat{r}$ for the target based on the extent to which each candidate explains the visual and dialogue context. We call it \textit{pragmatic inference} (see Figure~\hyperref[fig:prograsp]{2}). In line 7 at Algorithm~\hyperref[alg:prograsp]{1}, $\lambda$ is a rationality parameter~\cite{monroe2017colors,fried2018speaker,shen2019pragmatically} in the range $[0, 1]$ that indicates the relative importance of the evaluation from A-int in pragmatic inference.
\begin{algorithm}
\caption{Pragmatic Object Grasping}
\label{alg:prograsp}
\begin{algorithmic}[1]
\Require Modules for VG ($P_\mathcal{V}$), Q-gen ($P_\mathcal{Q}$), A-int ($P_\mathcal{A}$)
\Require Module for object grasping ($\mathcal{O}$)
\Require 2D RGB image $\mathcal{I}$ and dialogue history $\mathcal{D} \gets \left\{\ell\right\}$
\Require A human user to interact with the robotic system
\Require The empty set of object regions $\mathcal{R} = \emptyset$
\For{$t \gets 1$ to $T$}
    \State $\tilde{\mathcal{R}}_{t-1}$ $\gets$ $P_\mathcal{V}(\cdot | \mathcal{I}, \mathcal{D}_{\le t-1})$ where $\tilde{\mathcal{R}}_{t-1} = \left\{r_1, \cdots, r_{|\tilde{\mathcal{R}}|}\right\}$
    \State $\mathcal{R} \gets \mathcal{R} \cup \tilde{\mathcal{R}}_{t-1}$
    \State $\tilde{q}_t \gets P_\mathcal{Q}(\cdot | \mathcal{I}, \mathcal{D}_{\le t-1}, r^q_t)$ where $r^q_t \sim \mathcal{R}$
    \State The user provides an answer $\tilde{a}_t$ to the question $\tilde{q}_t$
    \State $\mathcal{D} \gets \mathcal{D} \cup \left\{\tilde{q}_{t}, \tilde{a}_{t}\right\}$
    \State $\hat{r} \gets \mathrm{argmax}_{r \in \mathcal{R}} \, P_\mathcal{A}(\tilde{a}_t | \mathcal{I}, r, \tilde{q}_t)^{\lambda} \cdot P_\mathcal{V}(r | \mathcal{I}, \mathcal{D}_{\le t})^{1-\lambda}$
\EndFor
\State Grasp the object $\mathcal{O}(\hat{r}, \mathcal{P})$ with $\hat{r}$ and the point cloud $\mathcal{P}$
\end{algorithmic}
\end{algorithm}
\vspace*{-1.0cm}
\subsection{Object Grasping Module}
The object grasping module (OG) first computes the 3D coordinates of the predicted object using the 2D region coordinates $\hat{r}$ and the point cloud $\mathcal{P}$. Specifically, OG matches the 2D object region with the point cloud on the identical resolution and then segments points inside the region. We employ the RANSAC~\cite{schnabel2007efficient} to remove the table plane from the segmented points. The 3D target coordinates are computed by averaging the segmented points. Finally, OG computes the motion planning~\cite{coleman2014reducing} and performs object grasping.

%% file: 04_experiment.tex
\section{EXPERIMENTS}

\subsection{Experimental Setup}
\begin{figure}[ht!]
\centering
\includegraphics[width=0.5\textwidth]{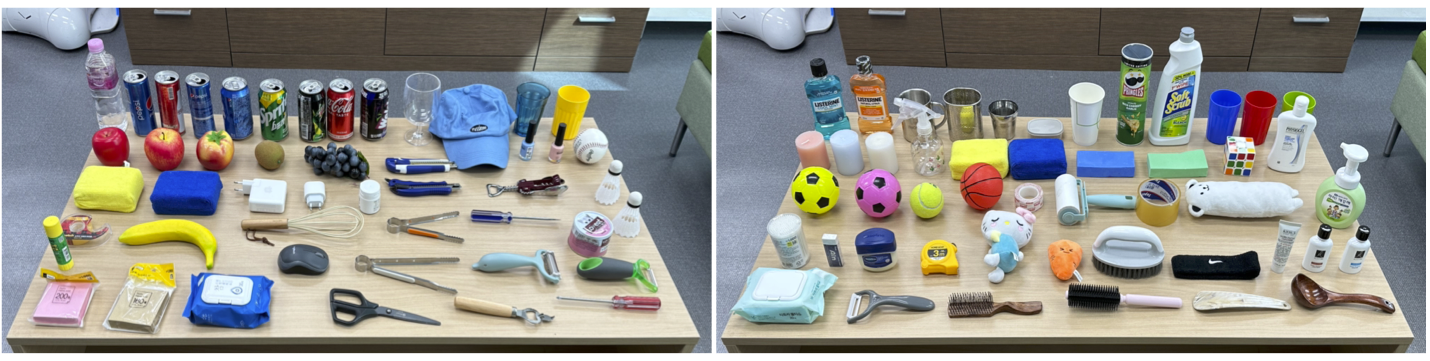}
\vspace*{-0.5cm}
\caption{\textbf{The 86 categories of everyday objects used in the experiments.}}
\label{fig:objects}
\vspace*{-0.4cm}
\end{figure}

\noindent\textbf{Intention-oriented Multi-modal Dialogue Dataset.} We evaluate our proposed method on the IM-Dial dataset, collected by the chatting between two players about images. The IM-Dial dataset consists of 800 images and 500 human-to-human dialogue data that cover 86 categories of everyday objects as shown in Figure~\hyperref[fig:objects]{3}. We divide the IM-Dial dataset into five splits: train, validation, test-seen, test-unseen, and test-cluttered. The train split contains 400 images and corresponding dialogue data for training. The validation split has 100 image and dialogue pairs. The test-seen, test-unseen, and test-cluttered data contain 100 pairs of images and intention-oriented utterances each. Note that these test splits do not have question-and-answer data, so the robotic system should identify the target object, interacting with a human user. The test-unseen split includes \textit{never-seen-before} objects not observed in the training procedure. The goal of the test-unseen split is to evaluate the generalization ability of the robotic system. Furthermore, we define the test-cluttered as a cluttered version of the test-seen split where objects are arbitrarily placed (e.g., (a) in Figure~\hyperref[fig:qualitative2]{6}). 

\noindent\textbf{Robotic Platform.} We conduct online experiments using a physical robot arm, the 6-DoF Kinova Gen3 lite with a two-fingered gripper. Our system utilizes Intel Realsense Depth Camera D435 to get an RGB-D image. The remote server processes our proposed algorithm and communicates with the robotic platform. The robotic platform locally computes motion trajectory planning.

\noindent\textbf{Baselines.} We compare PROGrasp with four methods:
\begin{itemize}
  \item \textbf{Zero-Shot}: The zero-shot approach is a visual grounding model not trained on the IM-Dial dataset. We implement it by finetuning the pre-trained OFA model on the visual grounding dataset, RefCOCO~\cite{yu2016modeling}. This approach predicts the object region given an input image and intention-oriented utterance (\textit{i.e.,} $\hat{r} = \mathrm{argmax}_{r \in \mathcal{R}} P_{\mathcal{V}}(r|\mathcal{I}, \mathcal{D}=\ell)$).   
  \item \textbf{SilentGrasp}: SilentGrasp is an ablative model of PROGrasp that does not have the question generation (Q-gen) and answer interpretation (A-int) modules. It predicts the object region in the same way as Zero-Shot, but the visual grounding module (VG) is trained on the IM-Dial dataset. 
  \item \textbf{LiteralGrasp}: LiteralGrasp is another ablative method of PROGrasp that does not have A-int. It is equivalent to PROGrasp without pragmatic inference (\textit{i.e.,} $\lambda=0$).
  \item \textbf{A-int-only}: This method is equivalent to PROGrasp that does not utilize VG in pragmatic inference (\textit{i.e.,} $\lambda=1$).
\end{itemize}

\subsection{Offline Experiments}

\noindent\textbf{Evaluation Protocol.} The offline experiment aims to verify how well the robotic system discovers the target object through human-robot natural language interaction. We measure the Intersection over Union (IoU) between the target object region $r^*$ and the predicted region after the interaction $\hat{r}$. The IoU is defined as the overlapping region between the two divided by their union region. The percentage of examples with an IoU value greater than 0.5 is typically reported as Acc$@0.5$. However, the threshold value of 0.5 may not be a reliable indicator to estimate the success of object grasping since object grasping requires accurate prediction of the target coordinates. We thus additionally report Acc$@0.9$, which requires more tight alignment between $r^*$ and $\hat{r}$.

\begin{table*}[ht!]
  \centering
  \caption{\textnormal{\textbf{Results on the offline experiments.} Underlined scores indicate the performance of the runner-up method.}}
  \resizebox{0.9\textwidth}{!}{
  \begin{tabular}{lccccccccc}
    \toprule
    \multirow{2}{*}{} & \multirow{2}{*}{} & \multicolumn{2}{c}{Validation} & \multicolumn{2}{c}{Test-Seen} & \multicolumn{2}{c}{Test-Unseen} & \multicolumn{2}{c}{Test-Cluttered} \\ 
    \cmidrule(lr){3-4}\cmidrule(lr){5-6}\cmidrule(lr){7-8}\cmidrule(lr){9-10} 
    Method & GDH & Acc$@0.5$ & Acc$@0.9$ & Acc$@0.5$ & Acc$@0.9$ & Acc$@0.5$ & Acc$@0.9$ & Acc$@0.5$ & Acc$@0.9$ \\
    \midrule
    Zero-Shot~\cite{wang2022ofa} & & 14\% & 4\% & 14\% & 7\% & 3\% & 2\% & 6\% & 4\%  \\
    SilentGrasp & & 50\% & 44\% & 54\% & 45\% & 45\% & 31\% & 41\% & 22\% \\
    A-int-only & & 82\% & \underline{75\%} & 81\% & 66\% & \underline{78\%} & \underline{57\%} & 83\% & 40\% \\
    LiteralGrasp & & \underline{84\%} & \underline{75\%} & \underline{85\%} & \underline{74\%} & 73\% & 54\% & \underline{84\%} & \underline{41\%} \\
    \midrule
    Zero-Shot~\cite{wang2022ofa} & $\checkmark$ & 51\% & 16\% & - & - & - & - & - & - \\
    SilentGrasp & $\checkmark$ & 83\% & 72\% & - & - & - & - & - & - \\
    \midrule
    \textbf{PROGrasp (ours)} & & \textbf{87\%} & \textbf{79\%} & \textbf{90\%} & \textbf{75\%} & \textbf{83\%} & \textbf{61\%} & \textbf{88\%} & \textbf{42\%} \\
    \bottomrule
  \end{tabular}}
  \label{tab:offline}
\end{table*}

\noindent\textbf{Results on the Validation Split.} We first evaluate PROGrasp and the compared methods on the validation split. As shown in Table~\hyperref[tab:offline]{1}, PROGrasp outperforms all compared methods on all evaluation metrics. It indicates our proposed components (\textit{i.e.,} Q-gen, A-int, and pragmatic inference) play a crucial role in boosting performance. Moreover, comparing Zero-Shot and SilentGrasp, even the strong pre-trained model performs poorly without adapting to Pragmatic-IOG.


We further study a new setting for Zero-Shot and SilentGrasp, called Grounding from Dialog History (GDH). We naturally assume Zero-Shot and SilentGrasp can only access the intention-oriented utterance (\textit{i.e.,} $\ell$) in the inference phase since neither approach has a module for question generation. However, the initial utterance is insufficient to pinpoint the target object. In GDH, we assume that Zero-Shot and SilentGrasp can access the ground-truth human-to-human dialogue history, so we feed the entire dialogue history (\textit{i.e.,} intention-oriented utterance and a set of question and answer pairs) into the models. As shown in Table~\hyperref[tab:offline]{1}, GDH significantly boosts Acc$@0.9$ of Zero-Shot (4\%$\rightarrow$16\%) and SilentGrasp (44\%$\rightarrow$72\%). The results indicate that additional question and answer pairs contain detailed information to identify the target object. Remarkably, PROGrasp outperforms SilentGrasp with GDH, although it does not require the ground-truth dialogue history. The results illustrate that PROGrasp works effectively, even in a more realistic scenario.

\noindent\textbf{Results on the Test Splits.} We compare PROGrasp with the compared methods on the test-seen, test-unseen, and test-cluttered splits. In Table~\hyperref[tab:offline]{1}, PROGrasp consistently yields improvements on all test splits compared with Zero-Shot, SilentGrasp, A-int-only, and LiteralGrasp. Specifically, compared with LiteralGrasp, PROGrasp improves Acc$@0.5$ on ten points (73\%$\rightarrow$83\%) and Acc$@0.9$ on seven points (54\%$\rightarrow$61\%) in the test-unseen split. We could not investigate the results of GDH on the test splits since the test splits do not have ground-truth dialogue data. Surprisingly, PROGrasp shows decent Acc$@0.5$ scores even in the test-cluttered and test-unseen splits, but relatively lower scores are observed on Acc$@0.9$. It illustrates that (1) accurately identifying partially occluded target objects and (2) generalizing a robotic system to previously unseen objects are challenging aspects of this task. We will discuss more details in the qualitative analysis.

\noindent\textbf{Comparison with the Multimodal Foundation Model.} We further identify the performance of the powerful multimodal foundation model, GPT-4V(ision)~\cite{gpt4v} on the validation split. GPT-4V is provided with images and detailed text prompts (see supplement). Table~\hyperref[tab:mfm]{2} shows the results. We observe that GPT-4V infers the target object well, but it poorly specifies the ground-truth region coordinates. We thus study a hybrid approach: (1) GPT-4V interacts with users and generates a distinctive caption of the best-fit object and (2) our VG model then predicts the coordinates $\hat{r}$ based on the caption. As shown in Table~\hyperref[tab:mfm]{2}, PROGrasp outperforms the approach on all evaluation metrics. It also demonstrates the significance of pragmatic inference.               

\begin{figure}[t!]
\centering
\label{fig:analysis}
\includegraphics[width=0.45\textwidth]{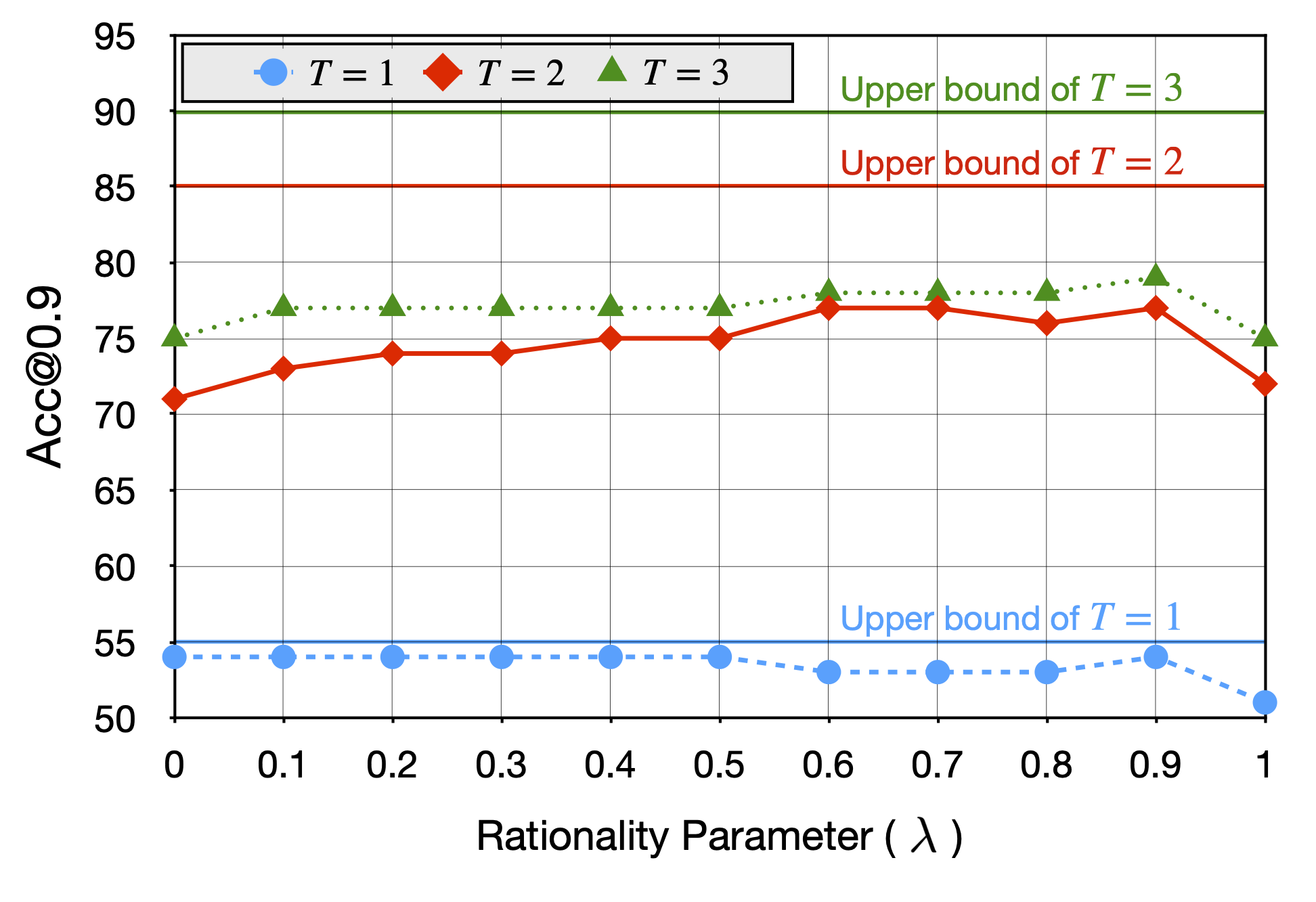}
\vspace*{-0.3cm}
\caption{\textbf{Validation scores adjusting the hyperparameters, $\lambda$ and $T$.}}
\vspace*{-0.5cm}
\end{figure}

\noindent\textbf{Hyperparameter Study.} We study how the hyperparameters in PROGrasp (\textit{i.e.,} $\lambda$ and $T$) affect performance. Note that $\lambda$ indicates the importance of the evaluation from A-int in pragmatic inference, and $T$ denotes the number of interactions between PROGrasp and the human in Algorithm~\hyperref[alg:prograsp]{1}. As shown in Figure~\hyperref[fig:analysis]{4}, we visualize the Acc$@0.9$ performance on the validation split. The blue, red, and green lines denote the results when $T=1$, $T=2$, and $T=3$, respectively. We observe a huge performance gap between $T=1$ and $T=2$. It implies that many incorrect guesses in the first round of dialogue are corrected in the second round. Moreover, comparing $T=2$ with $T=3$, improvements seem saturated. 

We also identify that performance varies depending on the value of the rationality parameter $\lambda$. LiteralGrasp is equivalent to $\lambda=0$, and A-int-only corresponds to $\lambda=1$. The best results are observed in $\lambda=0.9$ across all $T$ values. However, we do not see any merit in PROGrasp compared with LiteralGrasp when $T=1$. To delve into this phenomenon, we visualize the upper bound performance for each $T$ value. The upper bound is defined as the performance when the system perfectly selects the object region in a set of object region candidates $\mathcal{R}$. The upper bound performance of $T=1$ is 55\%, and PROGrasp and LiteralGrasp both show 54\%. It illustrates that there is little room for improvement in $T=1$. In contrast, the upper bound of $T=2$ is 85\%, and LiteralGrasp shows 71\% in Figure~\hyperref[fig:analysis]{4}. PROGrasp boosts 6\% compared with LiteralGrasp (71\%$\rightarrow$77\%). Likewise, we observe 4\% gains (75\%$\rightarrow$79\%) when $T=3$. Unless stated otherwise, $\lambda$ is 0.9, and $T$ is 3. 

\noindent\textbf{Communicative Efficiency.} Beyond task success, communicative efficiency is also an important criterion for pragmatic systems~\cite{fried2022pragmatics}. Accordingly, we measure the average number of interactions (\textit{i.e.,} question answering) required for the system to identify the target object. In this study, we assume that the dialogue immediately ends when the Intersection over Union (IoU) between the predicted and target regions is greater than 0.5. The efficiency can range from 1.0 to $T=3$. We compare PROGrasp with three baselines: Random, A-int-only, and LiteralGrasp. The Random randomly selects the predicted object in the set of candidates (\textit{i.e.,} $\mathcal{R}$ in Algorithm~\hyperref[alg:prograsp]{1}). In Table~\hyperref[tab:ce]{3}, PROGrasp consistently improves the efficiency across all test splits. It indicates that our system efficiently identifies the target object through fewer interactions.

\begin{table}[t!]
  \centering
  \caption{\textnormal{\textbf{Comparison with the Multimodal Foundation Model.}}}
  \resizebox{0.4\textwidth}{!}{
  \begin{tabular}{lccc}
    \toprule
    Method & Acc$@0.1$ & Acc$@0.5$ & Acc$@0.9$ \\
    \midrule
    GPT-4V~\cite{gpt4v} & 29\% & 9\% & 1\% \\
    GPT-4V~\cite{gpt4v} + VG & 82\% & 82\% & 68\%  \\
    \midrule
    \textbf{PROGrasp (ours)} & \textbf{87\%} & \textbf{87\%} & \textbf{79\%} \\
    \bottomrule
  \end{tabular}}
  \label{tab:mfm}
\end{table}
\begin{table}[t!]
  \centering
  \caption{\textnormal{\textbf{Analysis of Communicative Efficiency.} $\downarrow$ indicates lower is better.}}
  \resizebox{0.45\textwidth}{!}{
  \begin{tabular}{lccc}
    \toprule
    \multirow{2}{*}{} & \multicolumn{3}{c}{Avg. \# of Interactions $\downarrow$}\\ 
    \cmidrule(lr){2-4}
    Method & Test-Seen & Test-Unseen & Test-Cluttered \\
    \midrule
    Random & 1.76 & 2.00 & 1.98 \\
    A-int & 1.60 & 1.78 & 1.76 \\
    LiteralGrasp & 1.55 & 1.78 & 1.71 \\
    \midrule
    \textbf{PROGrasp (ours)} & \textbf{1.53} & \textbf{1.72} & \textbf{1.69} \\
    \bottomrule
  \end{tabular}}
  \label{tab:ce}
\end{table}
\begin{table}[t!]
  \centering
  \caption{\textnormal{\textbf{Results on the online experiments.}}}
  \resizebox{0.45\textwidth}{!}{
  \begin{tabular}{lccc}
    \toprule
    \multirow{2}{*}{} & Ambiguous & Non-Ambiguous & Total \\
    \cmidrule(lr){2-4}
    Method & \multicolumn{3}{c}{Object Discovery / Success Rate}\\ 
    \midrule
    SilentGrasp & 42 / 30 & 78 / 38 & 56 / 33 \\
    LiteralGrasp & 77 / 47 & \textbf{90} / \textbf{45} & 82 / 46 \\
    \midrule
    \textbf{PROGrasp (ours)} & \textbf{80} / \textbf{53} & \textbf{90} / \textbf{45} & \textbf{84} / \textbf{50} \\
    \bottomrule
  \end{tabular}}
  \label{tab:online}
  \vspace*{-0.5cm}
\end{table}

\subsection{Online Experiments}
\noindent\textbf{Evaluation Protocol.} We reproduce 100 images in the test-seen split and conduct online experiments to study how well the system picks the desired object up through human-robot dialogue. We compare PROGrasp with SilentGrasp and LiteralGrasp. The human rater evaluates object discovery and success rate. Object discovery measures whether the system correctly localizes the target. 

\noindent\textbf{Results.} We divide 100 samples into two categories: Ambiguous (60) and Non-Ambiguous (40). The Ambiguous is a set of samples that require disambiguating over two objects given an initial utterance. The Non-Ambiguous corresponds to the other samples. In Table~\hyperref[tab:online]{4}, PROGrasp achieves a total execution success rate of 50\%, outperforming all baselines. It demonstrates that the superiority of PROGrasp's target object discovery is successfully transferred to the success rate. Not surprisingly, PROGrasp is effective in the Ambiguous, boosting success rate of 23\% compared with SilentGrasp. However, we observe many failure cases, although the target object is correctly localized. The system often drops the objects during lifting or fails to grasp them since objects are highly unstructured (see Figure~\hyperref[fig:objects]{3}). More precise object grasping will mitigate this issue. We leave it as a future work.

\subsection{Qualitative Analysis}
\begin{figure}[t!]
\centering
\label{fig:qualitative}
\includegraphics[width=0.48\textwidth]{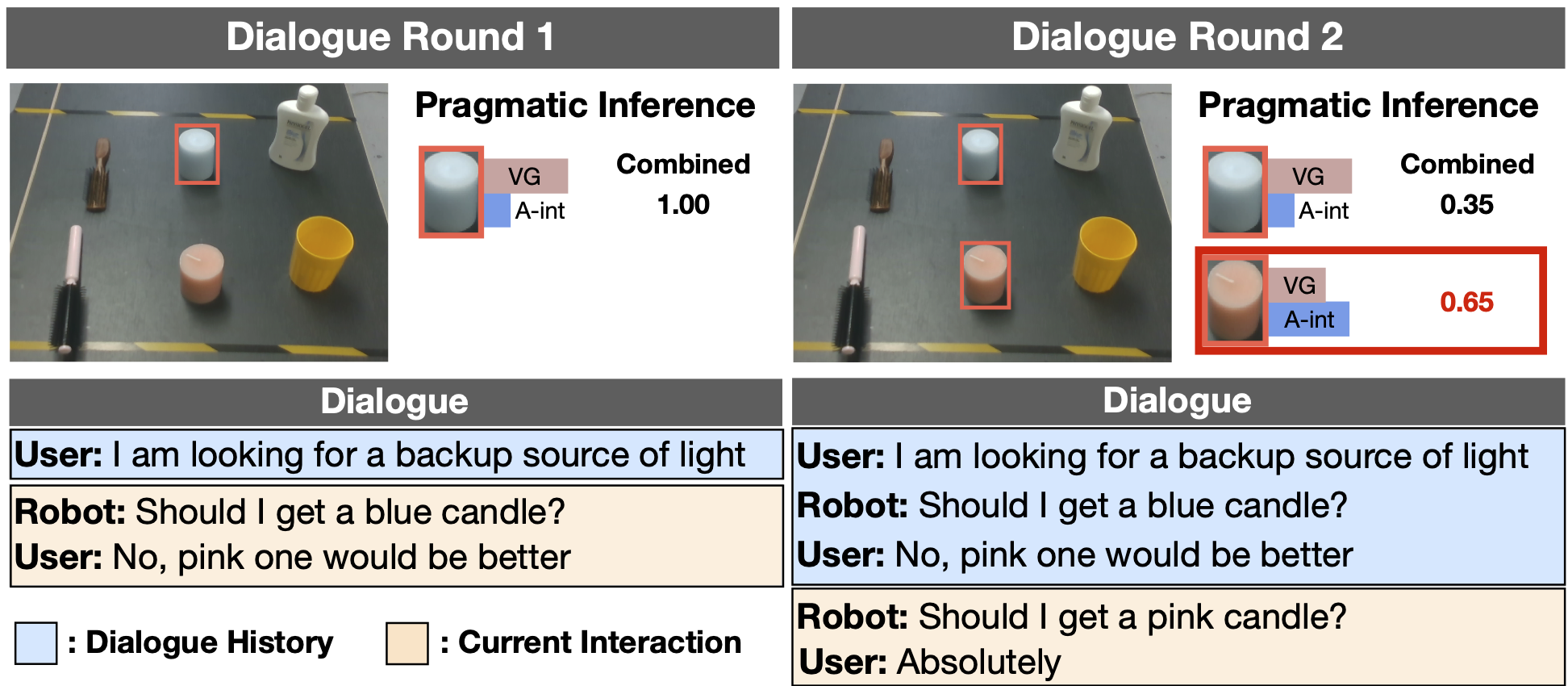}
\vspace*{-0.5cm}
\caption{\textbf{Visualization of PROGrasp's target object recovery.}}
\end{figure}

\begin{figure}[t!]
\centering
\label{fig:qualitative2}
\includegraphics[width=0.48\textwidth]{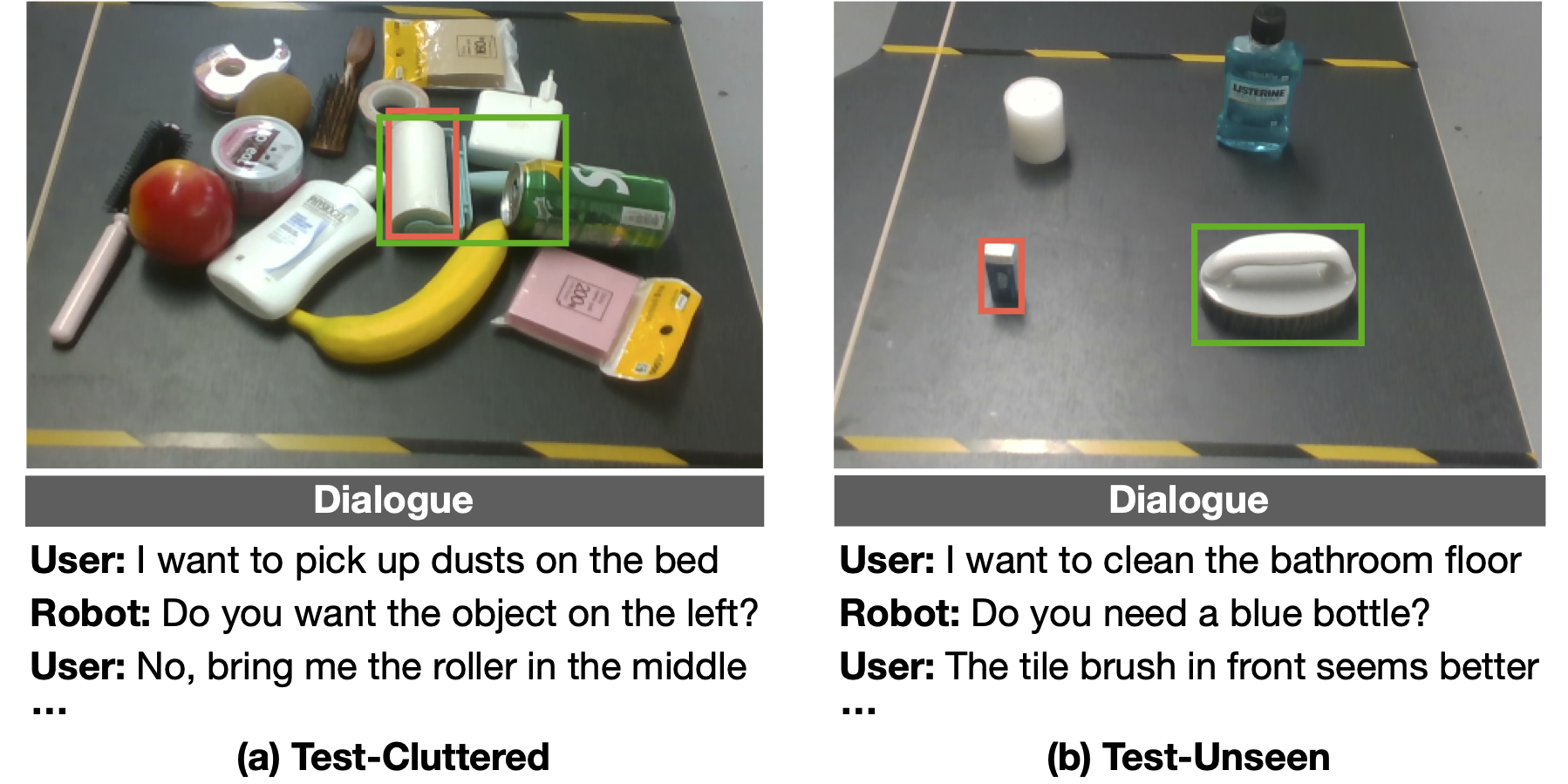}
\vspace*{-0.5cm}
\caption{\textbf{Visualization of the failure cases.}}
\vspace*{-0.5cm}
\end{figure}

\noindent\textbf{Results.} In Figure~\hyperref[fig:qualitative]{5}, we visualize the inferred results from PROGrasp when $T=2$. PROGrasp fails to find the target object (\textit{i.e.,} a pink candle) in the first round of the dialogue, but it corrects the target in the second round by utilizing the question-and-answer pair from the first round as additional context and includes the target object as a candidate. Pragmatic inference finally selects the desired object region. This example clearly explains the performance gap between $T=1$ and $T=2$ in Figure~\hyperref[fig:analysis]{4}. We also visualize two failure examples from the test-cluttered and test-unseen splits in Figure~\hyperref[fig:qualitative2]{6}. The red and green boxes indicate the predicted and ground-truth regions, respectively. As in (a) at Figure~\hyperref[fig:qualitative2]{6}, our system fails to identify the target region accurately due to the occlusion, which explains low scores in Acc$@0.9$ of the test-cluttered. Furthermore, PROGrasp sometimes makes an incorrect guess (\textit{i.e.,} (b) in Figure~\hyperref[fig:qualitative2]{6}) when observing never-seen-before objects, highlighting the need for further generalization in future work. 

%% file: 05_conclusion.tex
\section{CONCLUSION}
We present (1) a new task for interactive object grasping (Pragmatic-IOG) and (2) PROGrasp that effectively infers the user's intention for Pragmatic-IOG. Experiments demonstrate the effectiveness of our proposed methods, including pragmatic inference. We expect the Pragmatic-IOG to open the door to developing human-centric and pragmatic robots.

%% file: 06_appendix.tex
{\Large \noindent\textbf{Supplementary Materials}} 

\section{Intention-oriented Multi-modal Dialogue Dataset}

\noindent\textbf{Overview.} We propose the Intention-oriented Multi-modal Dialogue (IM-Dial) dataset to study the pragmatic reasoning behavior of the robotic system. The IM-Dial dataset consists of images and corresponding dialogues regarding 86 categories of everyday objects. The objects range from kitchen supplies (\textit{e.g.,} tongs, spoon, and sponge) to office supplies (\textit{e.g.,} memo pad, cutter, and tape). \\

\begin{figure*}[h!]
\centering
\label{fig:collection}
\includegraphics[width=0.93\textwidth]{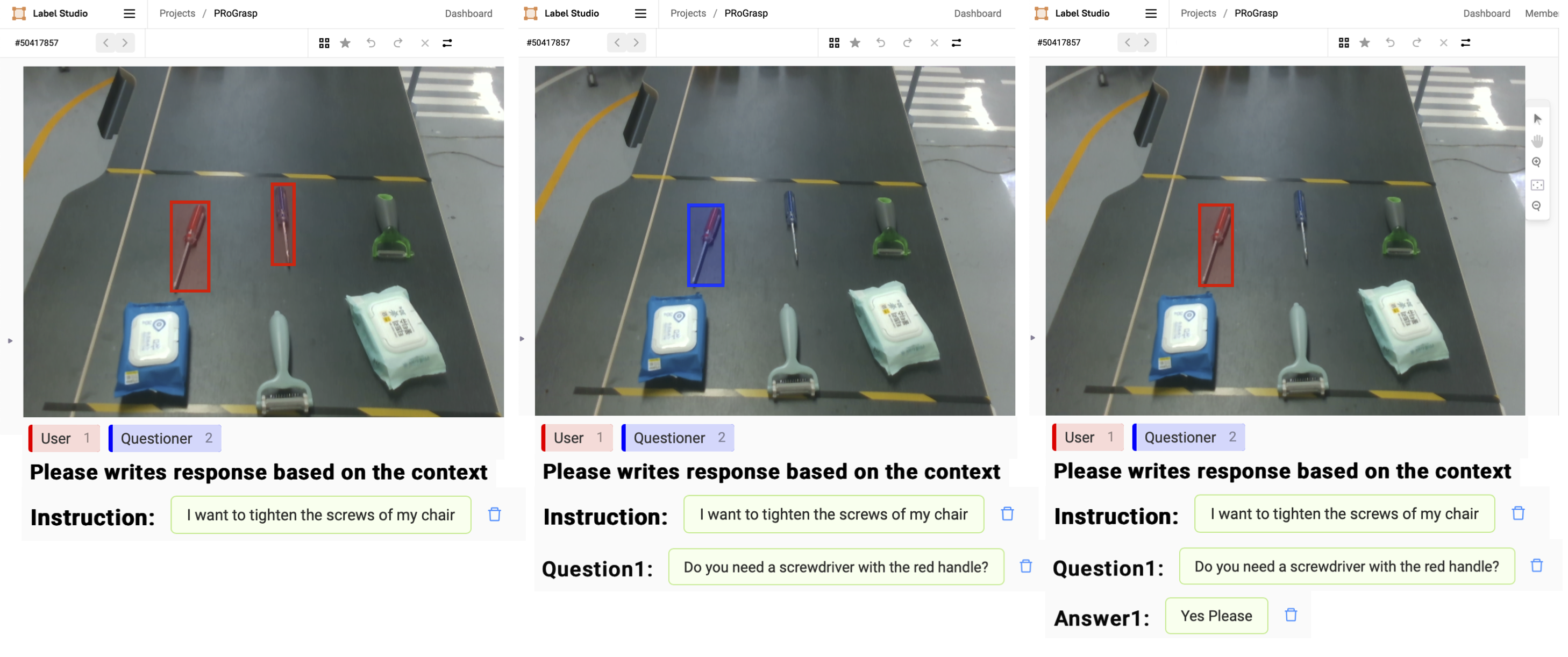}
\vspace*{-0.4cm}
\caption{\textnormal{\textbf{A visualization of a data collection tool.}}}
\end{figure*}

\noindent\textbf{Data Collection.} We collect the IM-Dial dataset using the labeling tool, Label Studio\footnote{https://labelstud.io}. Two players (\textit{i.e.,} the questioner and the answerer) log on to the online tool and cooperatively collect the IM-Dial dataset by chatting about images. As shown in Figure~\hyperref[fig:collection]{7}, the conversation begins with an intention-oriented utterance $\ell$ of the answerer, ``I want to tighten the screws of my chair.'' The answerer also annotates the candidate object regions (\textit{i.e.,} the red-colored bounding boxes; $\mathcal{R}_0$) that the visual grounding module (VG) should predict given the utterance. Next, the questioner annotates the most likely object region (\textit{i.e.,} the blue-colored bounding box; $r^q_1$) for the target object and asks a question $q_1$ to check whether the selected region corresponds to the target. The answerer provides the response $a_1$ to the question and annotates the region candidates $\mathcal{R}_1$ that VG should infer after seeing the response. Note that the questioner does not know the target object $r^*$, and the answerer can see which region the questioner annotates. The question answering can repeat until the questioner finds the target object. However, we observe that all dialogues have one pair of QA since human annotators already have pragmatic reasoning abilities and common sense. In other words, the questioner easily pinpoints the target region after seeing the response to the first question. We inspect the quality of the entire IM-Dial dataset and refine wrongly annotated data.

As a result, we collect 500 image-dialogue pairs and 300 image-instruction pairs. The image-dialogue pairs are divided into the training split (400 pairs) and the validation split (100 pairs). The image-instruction pairs are used as the test-seen, test-seen-cluttered, and test-unseen splits. The instruction denotes the intention-oriented utterance. In conclusion, in the test splits, the robotic system should interact with the human user to identify the target object. \\

\section{Implementation Details}

We finetune the $\mathrm{OFA_{Large}}$ model to train our proposed modules (\textit{i.e.,} VG, Q-gen, and A-int). We use the AdamW optimizer with $(\beta_1, \beta_2) = (0.9, 0.999)$ and $\epsilon=1e-8$ to train the modules. The learning rate is warmed up and linearly decays with a ratio of 0.01. We train each module on four NVIDIA RTX 3090 GPUs with a batch size of 1 for 10 epochs. It takes about a half hour to train each module. The size of the input image is 640 $\times$ 480. Q-gen and A-int can generate a sequence whose length is up to 16. The maximum sequence length of VG is 20. When training VG, the ground-truth region coordinates $r^*$ are quantized to integers ranging from 0 to 999. In other words, the input image is divided into 1000 $\times$ 1000 pieces, and VG predicts one of the values from 0 to 999 in each time step. \\

\section{Details Regarding Multimodal Foundations Models}

In Table~\hyperref[tab:mfm]{2}, we compare PROGrasp with the state-of-the-art multimodal foundation models, GPT-4V(ision)~\cite{gpt4v}. To make GPT-4V perform the Pragmatic-IOG task, we provide the model with detailed text prompts. The prompts consist of task descriptions, output format, and examples. The task description contains the details about the task. The output format specifies the format of the desired output. For example, in the case where GPT-4V should predict the target object's coordinates, the output format has the description of the coordinate system and the desired output. The example part contains what GPT-4V should return in a certain scenario. We visualize the specific text prompts of GPT-4V in the following page.

\onecolumn
\clearpage

\noindent\textbf{1. Default Prompt for GPT-4V}

\begin{tcolorbox}[breakable,boxrule=0pt,width=\textwidth]
  You are a generalist agent following natural language instructions through dialogue with humans. Please return the desired output by referring to the following explanation. \\

  TASK DESCRIPTION: \\
  The task you should perform requires pragmatic reasoning capabilities. Specifically, your goal is to localize the target object in the given image based on natural language instructions (i.e., visual grounding), but the instructions do not explicitly contain the category of the target object. The instructions are intention-oriented, for example, "I am thirsty" or "My device runs out of battery." So, you need to infer the intent of the instructions based on context (i.e., an input image) and localize one target object that best matches the given instruction. \\

  Furthermore, the instructions are often ambiguous. For example, you can see more than two drinkable objects in the image when the initial instruction is "I am thirsty." In this case, you can ask questions to disambiguate the target object, such as "Do you need the can of Coke?". The human users will answer your questions. You can ask up to three questions for each sample. If you think the target object is clearly specified, you should finally generate a distinctive description of the target object. \\

  OUTPUT FORMAT: \\
  Your output should be either the question for disambiguation or the bounding box coordinates of the target object. You can ask questions in the middle of the dialogue, but the final output should be the bounding box coordinates. The bounding box coordinates should follow the format: [x1, y1, x2, y2]. In other words, you should return the list whose length is four, and each entry should be the Python float type. x1 and y1 denote the bounding box's top left xy coordinates. x2 and y2 denote the bounding box's bottom right xy coordinates. The top left corner of the given image is (0, 0), and the input image size is 640 x 480. So, the x value should be between 0 and 640, and y should be between 0 and 480. Based on the definition of coordinates, x2 should be larger than x1, and y2 should be larger than y1. Note that the bounding boxes should enclose the target object tightly. \\

  EXAMPLE: \\
  The instruction is "I am hungry". You detect kiwi, strawberry, tomato, and banana in the image. So, you can ask, "Do you need the banana?". The human user answers "No." So, you ask the follow-up question, "Do you want to eat the strawberry?" Then, the human says, "No, the kiwi seems better for me." By inferring the meaning of the initial instruction and dialogue history (a series of question and answer pairs), you should return the bounding box coordinates of the kiwi in the image. If you see the kiwi near the bottom right corner of the image, the coordinates, for example, can be [602.31, 450.53, 635.72, 475.00]. \\

  Based on the task description, output format, and example above, please localize the target object through dialogue with humans. The current instruction is $\left\{\text{Instruction}\right\}$. 
\end{tcolorbox}

\clearpage
\noindent\textbf{2. Prompt for GPT-4V in the Hybrid Approach (GPT-4V + VG)}
\begin{tcolorbox}[breakable,boxrule=0pt,width=\textwidth]
  You are a generalist agent following natural language instructions through dialogue with humans. Please return the desired output by referring to the following explanation. \\

  TASK DESCRIPTION: \\
  The task you should perform requires pragmatic reasoning capabilities. Specifically, your goal is to specify a target object in the given image based on natural language instructions, but the instructions do not explicitly contain the category of the target object. The instructions are intention-oriented, for example, "I am thirsty" or "My device runs out of battery." So, you need to infer the intent of the instructions based on context (i.e., an input image) and specify one target object that best matches the given instruction. \\

  Furthermore, the instructions are often ambiguous. For example, you can see more than two drinkable objects in the image when the initial instruction is "I am thirsty." In this case, you can ask questions to disambiguate the target object, such as "Do you need the can of Coke?". The human users will answer your questions. You can ask up to three questions for each sample. If you think the target object is clearly specified, you should finally generate a distinctive description of the target object. \\

  OUTPUT FORMAT: \\
  Your output should be either the question for disambiguation or the final description of the target object. Both should be generated in one sentence. \\
 
  EXAMPLE: \\
  The instruction is "I am hungry". You detect kiwi, strawberry, tomato, and banana in the image. So, you can ask, "Do you need the banana?". The human user answers "No." So, you ask the follow-up question, "Do you want to eat the strawberry?" Then, the human says, "No, the kiwi seems better for me." You should describe the target object, such as "The kiwi." \\ 

  Based on the task description and example above, please describe the target object through dialogue with humans.
  The current instruction is $\left\{\text{Instruction}\right\}$.  
\end{tcolorbox}

\label{prompt1}